\documentclass[final,11pt,times]{elsarticle}
\usepackage{xcolor}
\usepackage{float}
\usepackage{amsmath,amssymb,mathtools}
\usepackage{graphicx}
\usepackage{subcaption}
\usepackage[hidelinks]{hyperref}
\usepackage{relsize}
\usepackage{cleveref}
\crefname{figure}{Fig.}{Figs.}
\usepackage{booktabs}
\usepackage{threeparttable}
\usepackage{hyperref}
\usepackage{titlesec}
\usepackage{times}
\usepackage{etoolbox} % For patching commands

\makeatletter
\renewcommand{\subsection}{%
  \@startsection{subsection}{2}{\z@}%
    {-2ex \@plus -1ex \@minus -.2ex}%
    {0.5ex \@plus .2ex}%
    {\normalfont\bfseries} % Title font style
    [] % Bold section number
}
\makeatother

\usepackage{amssymb}

\usepackage{geometry}
\title{ Local vs. Global Models for Hierarchical Forecasting}

\author{\textbf{ZHAO YINGJIE}\corref{cor1}\fnref{label1}}
\author{\textbf{MAHDI ABOLGHASEMI}\corref{cor1}\fnref{label2}}

\affiliation[label1]{organization={Department of Data Science,\\ The University of Queensland},
            state={QLD 4702},
            country={Australia},
            email={\\ $yingjie.zhao@student.uq.edu.au$}}
\par\vspace{1cm}
\affiliation[label2]{organization={School of Mathematical Sciences, \\Queensland University of Technology},
            state={QLD},
            country={Australia},
            email={\\$mahdi.abolghasemi@qut.edu.au$}}

\cortext[cor1]{Corresponding author}

\begin{document}
\begin{frontmatter}

\begin{abstract}
Hierarchical time series forecasting plays a crucial role in decision-making in various domains while presenting significant challenges for modelling as they involve multiple levels of aggregation, constraints, and availability of information. This study explores the influence of distinct information utilisation on the accuracy of hierarchical forecasts, proposing and evaluating locals and a range of Global Forecasting Models (GFMs). In contrast to local models, which forecast each series independently, we develop GFMs to exploit cross-series and cross-hierarchies information, improving both forecasting performance and computational efficiency. We employ reconciliation methods to ensure coherency in forecasts and use the Mean Absolute Scaled Error (MASE) and Multiple Comparisons with the Best (MCB) tests to assess statistical significance. The findings indicate that GFMs possess significant advantages for hierarchical forecasting, providing more accurate and computationally efficient solutions across different levels in a hierarchy. Two specific GFMs based on LightGBM are introduced, demonstrating superior accuracy and lower model complexity than their counterpart local models and conventional methods such as Exponential Smoothing (ES) and Autoregressive Integrated Moving Average (ARIMA).
\end{abstract}

\begin{keyword}
Global Models \sep Hierarchical Time Series \sep Forecasting \sep Machine Learning
\end{keyword}

\end{frontmatter}

%\thispagestyle{fancy}


\section{INTRODUCTION}
\label{sec:introduction}
% background of HF and the traditional method: local models and its limitations
\noindent Time series forecasting plays a crucial role in decision-making across various domains in energy \citep{english2024improving}, demand forecasting \citep{abolghasemi2023value}, inventory control \citep{abolghasemi2020demand}, and finance \citep{SEZER2020106181}. Hierarchical time series refers to a collection of time series with constraints, in which time series always aggregates or disaggregates across levels with respect to some metric, such as geographical locations. Hierarchical forecasting can be challenging due to the complexity of the structure and relationships among the collection of series. \\

\noindent The traditional forecasting approach for hierarchical time series involves developing a local model for each time series, with the number of local models determined by the number of time series, and then reconciling them to generate consistent forecasts across different levels. This may result in expensive computations. The computational expense will exponentially increase as the growth in data size, and the model complexity will increase dramatically as the number of models increases. Although the cross-sectional information is shared between hierarchical series through reconciliation, the local models do not learn from the information in a hierarchy, i.e., local models do not exploit cross-sectional information from other time series. This can result in inefficient use of information from the hierarchy and missing out on the benefits of cross-sectional data when creating base forecasts.\\

% compared with local models to introduce GFMs and their characteristics
\noindent  The so-called Global Forecasting Models (GFM) can estimate the parameters of a large forecasting model by utilising information from multiple time series at once \cite{januschowski2020criteria}. The complexity of GFMs will remain as the data size increases and GFMs can be multivariate rather than limited to univariate in local models. These characteristics ensure that GFMs possess the feasibility and the ability to achieve accurate forecasts by cross-learning different information in hierarchies and among time series.\\

% Introduce an important factor in model performance (the subject in our paper) information utilisation
\noindent One crucial factor that significantly affects the performance of the global models is the type of information used during the training phase. Appropriate information utilisation can help models generate more accurate forecasts and generalise to future unseen values \citep{abolghasemi2023value}. On the other hand, under-utilisation or over-utilisation of information may also lead to problems with under-fitting or over-fitting. \citet{MONTEROMANSO20211632} affirm individual time series cannot be data-driven modelling, which leads to the over-fitting of local models by the sample size. Local models are determined solely by the individual time series, but GFMs consider multiple time series from one or more hierarchies to extract more complex relationships. As such, the optimal choice of information utilisation during the training phase is crucial for GFMs. The aim of this study is to investigate the effect of distinct information utilisation in developing GFM for hierarchical time series, thus determining the most effective methods for handling time series data at different levels.\\

% expound on the works in our paper
\noindent This research empirically investigates the hierarchical time series for 3049 products from Walmart as used in the M5 forecasting competition \citep{makridakis2021m5}. We develop several different models including local and global models by utilising different information across different hierarchical levels of particular products, and across different hierarchies of products. Several benchmarking and advanced machine learning based models are employed, and reconciliation methods, namely Bottom-Up (BU), Top-Down (TD) and Minimum Trace (MinT), are implemented to maintain consistency and coherency of forecasts.\\

\noindent The rest of this paper is structured as follows: Section \ref{sec:literature review} provides a brief literature review, while Section \ref{sec:method} explains the methodology used. Section \ref{sec:setup} outlines the experimental setup, and Section \ref{sec:results} presents the empirical results. Lastly, Section \ref{sec:conclusion} concludes the study.

\section{LITERATURE REVIEW}
\label{sec:literature review}

\noindent Local models have been demonstrated to be empirically effective in time series forecasting including hierarchical time series forecasting \cite{english2024improving}. Many statistical and machine learning models have been developed and tested for countless ranges of forecasting problems with somewhat contradicting results from one setting to another, affirming the performance of models depends on the availability of information, forecasting horizon, and time series behaviour such as intermittency and seasonality, among others \citet{abolghasemi2022machine,duncan2001forecasting}.   \citet{makridakis2018statistical} showed that statistical methods, particularly ARIMA and theta, demonstrate superior performance to the models across a range of forecast horizons. \citet{MEJIA2021107371} constructed a neural network combined by convolutional and pooling layers of Convolutional Neural Network (CNN) and units of Long-short Term memory (LSTM), which the proposed model outperforms ARIMA, Neural Network (NN). While there is evidence that local models can have satisfactory performance in both individual time series forecasting and hierarchical forecasting settings, there has been little research to investigate whether global models can be effective as they use information across series and levels.\\

 The early versions of GFMs were proposed by \citet{kourentzes2014improving}, who extract patterns in a class of time series and explore the influence on forecasts caused by differences and similarities between time series. \citet{smyl2019hybrid} also proposed a GFM, ES-RNN, which is the winning solution of the M4 competition and has been observed to perform well on yearly, quarterly and monthly data but still poses issues. The model shows poor performance on daily and weekly data and its computational cost is inestimable as the weights of patterns must be manually intervened when the number of time series is large. Another is the high complexity of the data preprocessing and model structure due to a multitude of parameters. \citet{NYSTRUP20211127} and \citet{trapero2015identification} provide a perspective that dimensionality reduction improves forecast accuracy. Based on M5 competition data \citep{makridakis2021m5}, \citet{BANDARA20221400} proposed a fast GFM capable of handling long memory data, achieving top 50 performance in M5 competition. The ensemble model comprises LightGBM and Pooling Regression, which combine two forecasts by arithmetic means. \citet{MONTEROMANSO20211632} examine the generalization of GFMs based on datasets including M1, M3, M4, Tourism, and others, finding GFMs without any preset assumptions achieve high accuracy. Furthermore, linear GFMs demonstrate competitive accuracy with a significantly reduced number of parameters compared to local models and the complexity of GFMs remains unlike the growth of local model complexity as the set size increases.

\section{DATA}
\label{sec:data}
%\captionsetup[figure]{labelformat=simple, labelsep=colon, justification=centering, font=small, labelfont=bf, textfont=bf, format=plain, singlelinecheck=off, skip=11pt}
\begin{figure}
    \centering
    \includegraphics[width=0.89\linewidth]{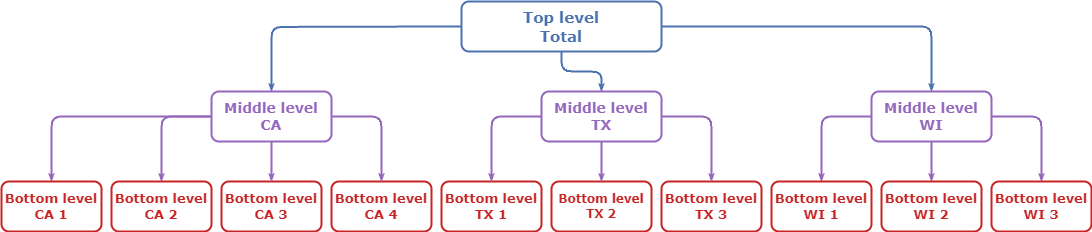}
    \caption{\textbf{Structure of Hierarchy}}
    \label{fig: Fig.1}
\end{figure}

\noindent The M5 competition dataset \citep{makridakis2021m5}, which encompasses sales, selling price holidays and promotional periods for 3,049 products over five years, containing 42,840 hierarchically organised time series over 12 levels. We select all the series across levels 10, 11, and 12 as the primary focus of this study. The 3049 products correspond to 3049 hierarchies, and the 14 time series are organised hierarchically across levels: the total, state and store levels. \autoref{fig: Fig.1} intuitively presents a hierarchy structure where each square represents one time series. Each time series comprises 1,941 sales data points and was formatted into an embedded matrix with a train-test split. The initial rolling window extracts 60 lag sales and a single sale of the day. The 61 data points are subsequently considered explanatory variables, and a new rolling window is employed to generate a single future sale as the dependent variable. Each row represents a small time series containing 62 sales, corresponding to 61 past and 1 future. The final 28 days' sales were considered the test set, while the remaining sales were used as the training set. Consequently, the scale of the data is 3049 $\times$ 14 $\times$ 1880 = 80,249,680, given that the sizes are 60 and 1 for the two rolling windows. 

\section{METHODOLOGY}\label{sec:method}
\subsection{\normalfont\textbf{Information Utilisation}}
\noindent Information utilisation is a non-negligible factor that could impact the accuracy of hierarchical forecasting models, as mentioned in \autoref{sec:introduction}. Therefore, it is essential to ascertain which type of information for cross-learning is included in GFMs. In this study, information utilisation are categorised into three groups: (i) information of a single series, (ii) cross-series information within the same hierarchy, and (iii) cross-hierarchical information. The distinct information utilisation will yield different effects on forecasting by dominating the training phase for models.

\subsection{\normalfont\textbf{Forecasting Models}}
\noindent Exponential Smoothing (ES) was proposed by \citet{HOLT20045}, generating the forecasts by giving weights to averages of lag observations. The simple ES in this study is one of the benchmarks and its mathematical formulation is $\hat{y}_{t+1} = \alpha y_t +(1-\alpha)\hat{y}_t$ where $y_t$ is the observation on time $t$, $\hat{y}_{t+1}$ is the forecast on time $t+1$, and $\alpha$ is the weight, $0 \leq \alpha \leq 1$. The larger $\alpha$ means that we allocate larger weights to values of recent time steps, and as the observations become older and older, the weights exponentially decay.\\

\noindent Autoregressive Integrated Moving Average (ARIMA) is a typical statistical learning model that integrates autoregressive (AR) and moving average (MA) components, and these techniques correspond to parameters $p$, and $q$. ARIMA achieves forecasting by observing the autoregressive component of past values, applying $d$-order difference to achieve stationarity, and using moving averages $q$ of order to adjust residuals.\\

\noindent LightGBM (LGBM) is a powerful and efficient algorithm proposed by \citet{ke2017lightgbm} that is fast and flexible. The leaf-wise growth strategy makes it fast and suitable for handling large volumes of data. Histogram-based optimisation speeds up feature split computation and reduces the time required to find split points by discretising feature values into bins and parallelising training.

\subsection{\normalfont\textbf{Reconciliation}}
\noindent Reconciliation methods are introduced to guarantee the consistency of the base forecasts for hierarchical time series. We apply three different models including TD, BU, and MinT. The summing matrix \textbf{\textit{S}} is the same for the three methods, showing the relationship between time series across levels within a hierarchy. Assume the number of time series in the hierarchy is $N_h$, and the number of time series at the bottom level is $N_b$, the formulations of the summing matrix and the relationship are $\textbf{\textit{S}}_{N_h \times N_b}$ and $\textbf{\textit{y}}_t = \textbf{\textit{S}}\textbf{\textit{b}}_t$. The variable $\textbf{\textit{y}}_t$ represents the time series in the hierarchy, and $\textbf{\textit{b}}_t$ represents the time series at the bottom level. Moreover, mapping matrix \textbf{\textit{SG}} determines how the base forecasts are mapped to coherent forecasts. We could receive forecasts from models, called base forecasts, and denote them by $\hat{y}_h$, where the subscript $h$ represents the forecast horizon. The coherent forecasts $\Tilde{y}_h$ can then be represented as $\Tilde{y}_h = \textbf{\textit{SG}}\hat{y}_h$, where the matrix \textbf{\textit{G}} is used to map the base forecasts into the bottom level and the distinct \textbf{\textit{G}} matrices will be used in below methods.\\

\noindent The Bottom-Up (BU) is straightforward and produces coherent forecasts based on the base forecasts at the bottom level. One advantage of BU is forecasting at the lowest level of the hierarchy to respect the constraints of the data to the greatest extent possible due to aggregation. The $\textbf{\textit{G}}_{\text{BU}}$ in the hierarchy will be identical for all products and is depicted as $\textbf{\textit{G}}_{\text{BU}} = \begin{bmatrix}
    \textbf{0}_{10\times4} & \textbf{\textit{I}}_{10\times10} 
\end{bmatrix}$.\\

\noindent The Top-Down (TD) generates coherent forecasts based on the base forecasts at the top level and will be allocated to forecasts at the lower level according to their proportions, whose proportions are estimated in several computational ways. We assumed the proportions of the historical averages as proportion estimation. The following mathematical notations of the proportions of the historical average and matrix $\textbf{\textit{G}}_\text{TD}$ are $\textbf{\textit{p}}_j = \frac{\sum^{T}_{t=1} y_{j,t} / T}{\sum^{T}_{t=1} y_{t} / T}, j = 1,...,m;$ and 
$\textbf{\textit{G}}_{\text{TD}} = \begin{bmatrix}
    \textbf{\textit{p}}_{10\times1} & \textbf{0}_{10\times13} 
\end{bmatrix}$.\\

\noindent The Minimum Trajectory method (MinT) differs from the previous two methods in the way it generates the matrix $\textbf{\textit{G}}_\text{MinT}$. The formulations of $\textbf{\textit{G}}_\text{MinT}$ and coherent forecasts $\Tilde{y}_{h,\text{MinT}}$ are $\textbf{\textit{G}}_{\text{MinT}} = 
(\textbf{\textit{S}}{'} \textbf{\textit{W}}^{-1}_{h} \textbf{\textit{S}})^{-1} \textbf{\textit{S}}' \textbf{\textit{W}}^{-1}_{h} \text{ and }
\Tilde{y}_{h,\text{MinT}} = 
(\textbf{\textit{S}}{'} \textbf{\textit{W}}^{-1}_{h} \textbf{\textit{S}})^{-1} \textbf{\textit{S}}' \textbf{\textit{W}}^{-1}_{h} \hat{y}_{h}.$ In order to obtain the estimation matrix $\textbf{\textit{W}}_h$ to generate the matrix $\textbf{\textit{G}}_\text{MinT}$, specific assumptions must be made in MinT.  We posit that the variance of each base forecast error on the bottom level is equal to $k_h$, where $k_h$ is a positive constant and is uncorrelated between nodes. This is expressed as $\textbf{\textit{W}}_{h} = k_{h}  \boldsymbol{\Lambda} $, where $k_h > 0$ and $ \boldsymbol{\Lambda} = \text{diag}(\textbf{\textit{S}1})$, and $\textbf{1}$ is a unit vector of dimension $m$ (the number of time series at the bottom level).

\section{EXPERIMENTAL SETUP}\label{sec:setup}
\noindent The benchmarking methods, ES and ARIMA, were considered primary benchmarks for comparison with the LightGBM. Each row developed the ES and ARIMA models in the embedding matrix, where each row represents a small time series with a length of 62. Moreover, given the constraints imposed by the length of each row, ES and ARIMA have no seasonality or trend involved; thus, the default parameters are employed to construct these models. We constructed the Simple ES and ARIMA models from the $statsmodels$ library in Python 3.11.1.\\

\noindent 
We also developed LGBM models and compared them with the benchmarks. LGBM models were categorised into two groups which were trained by distinct information utilisation: (i) local LGBM (loc\_LGBM); (ii) the not fully global LGBM and the fully global LGBM (nfg\_LGBM and fg\_LGBM). On the one hand, the loc\_LGBM was developed based on the embedding matrix for a single time series. Namely, each loc\_LGBM was developed by a data matrix with 1880 rows and 62 columns, and there are 3049 $\times$ 14 = 42,686 loc\_LGBM. We assumed the boosting type and distribution were the Gradient Boosting Decision Tree (GBDT) and Tweedie. Tweedie facilitated the model's ability to learn and utilise the uniformity of occurrence. The learning rate is set between 0.01 and 0.11, with an increment step of 0.02, and the feature fraction is set between 0.3 and 0.7, with an increment step of 0.2. On the other hand, GFMs were trained based on two information utilisation strategies. nfg\_LGBM exploits cross-series information, and fg\_LGBM utilises cross-hierarchical information. It should be noted that the parameter setting for the two GFMs is identical to that of the local LightGBM. Besides, 3049 nfg\_LGBMs were developed for 3049 products (3049 hierarchies), but only one fg\_LGBM was developed for 3049 products since fg\_LGBM was designed to process all hierarchies. All LGBM models were based on $lightgbm$ in Python 3.11.1.

\section{EMPIRICAL RESULTS}\label{sec:results}
\noindent In total, $\text{3049 }(hierarchies)\times\text{14 }(time\text{ }series)\times \text{28 }(testset\text{ }size) = \text{1,195,208}$ forecasts for each model were evaluated across five evaluated levels. We take a simple average to compute the average accuracies across all levels and products. The mathematical formulations are as follows,
\begin{align}
\text{Avg-Levels} =  \frac{1}{3}(\sum^{3}_{j=1} \frac{1}{3049}(\sum^{3049}_{i=1} \frac{1}{N_j} (\sum^{N_j}_{n=1} \text{MASE}^{n,j}_i))),\\
\text{Avg-Products} = \frac{1}{3049}(\sum^{3049}_{i=1} \frac{1}{14}(\sum^{N=14}_{n=1} \text{MASE}^{n}_{i})),
\end{align}

\noindent where the $j$ represents the level, and $N$ and $N_j$ denote the number of time series in the level $j$. \citet{abolghasemi2022machine} utilised Avg-Ls as a standard to evaluate model performance, taking model performances on average of the three levels into account. The Avg-Ps are proposed to evaluate models' generalisation, stability and reliability,  considering the model's average performance on products. This is plausible since decision-making could be optimised solidly if the specific model could outperform others in different metrics.

\subsection{\normalfont\textbf{Performances and Stability Evaluation}}

%\captionsetup[table]{labelformat=simple, labelsep=colon, justification=centering, font=small, labelfont=bf, textfont=bf, format=plain, singlelinecheck=off, skip=11pt}
\begin{table}
\centering
\begin{threeparttable}
\caption{MASE for forecasting hierarchical methods across 3049 products at each level.}
\label{tab:1}
\begin{tabular}{lccccc}
\toprule
\textbf{Models} & \textbf{Top level} & \textbf{Middle level} & \textbf{Bottom level} & \textbf{AvgLevels} & \textbf{AvgProducts} \\
\midrule
ES & 2.6054 & 1.2055 & 1.0412 & 1.6174 & 1.0233 \\
ES-BU & 2.4985 & 1.1887 & 1.0412 & 1.5761 & 1.0188 \\
ES-TD & 2.6054 & 1.2391 & 1.0915 & 1.6453 & 1.0517 \\
ES-MinT & 2.5517 & 1.1999 & 1.0477 & 1.5998 & 1.0240 \\
ARIMA & 1.3239 & 0.9265 & 0.9383 & 1.0629 & 0.9224 \\
ARIMA-BU & 1.3218 & 0.9249 & 0.9383 & 1.0617 & 0.9219 \\
ARIMA-TD & 1.3239 & 0.9640 & 1.0220 & 1.1033 & 0.9775 \\
ARIMA-MinT & 1.3229 & 0.9261 & 0.9408 & 1.0633 & 0.9246 \\
loc\_LGBM & 1.3304 & 0.9684 & 0.8654 & 1.0547 & 0.8697 \\
loc\_LGBM-BU & 1.5190 & 1.0434 & 0.8654 & 1.1426 & 0.8968 \\
loc\_LGBM-TD & 1.3304 & 0.9810 & 0.8995 & 1.0703 & 0.9029 \\
loc\_LGBM-MinT & 1.3819 & 0.9980 & 0.9204 & 1.1001 & 0.9135 \\
nfg\_LGBM & 1.1596 & 0.8120 & \textbf{0.8235} & 0.9317 & 0.8418 \\
nfg\_LGBM-BU & 0.8686 & 0.8089 & \textbf{0.8235} & 0.8337 & \textbf{0.8261} \\
nfg\_LGBM-TD & 1.1596 & 0.9291 & 0.8927 & 0.9938 & 0.8862 \\
nfg\_LGBM-MinT & 0.9177 & 0.8232 & 0.8291 & 0.8567 & 0.8299 \\
fg\_LGBM & \textbf{0.7732} & \textbf{0.8060} & 0.9072 & 0.8288 & 0.8970 \\
fg\_LGBM-BU & 0.7991 & 0.8184 & 0.9072 & 0.8416 & 0.9010 \\
fg\_LGBM-TD & \textbf{0.7732} & 0.8500 & 1.0046 & 0.8759 & 0.9528 \\
fg\_LGBM-MinT & \textbf{0.7732} & 0.8062 & 0.8967 & \textbf{0.8254} & 0.8872 \\
\bottomrule
\end{tabular}
\end{threeparttable}
\end{table}

% Local models
\noindent \autoref{tab:1} demonstrates that models' performances differ significantly across the five evaluated levels. In the local models, the performances generated by loc\_LGBM and ARIMA are very similar at the top level, which were 1.3304 and 1.3239, respectively. For the local LightGBM, its performance of  0.9684 is poorer than the 0.9265 of ARIMA at the middle level, but at the bottom level, its performance of 0.8654 is slightly better than 0.9383. Regarding the average for levels, the loc\_LGBM got 1.0547, close to 1.0629 of ARIMA. At the average of hierarchies (products), the loc\_LGBM with 0.8697 MASE is better than the 0.9224 of ARIMA. Nevertheless, the ES performed the worst compared with the loc\_LGBM and ARIMA across the five evaluated levels, with the worst performance being 2.6054 for the top level, which could signal that the loc\_LGBM easier to capture the trend of time series at the bottom level. Based on the above comparisons, LGBM has no significant advantages over ARIMA. We could observe the effect on forecasting caused by the cross-learning and information utilisation involved in GFMs.\\

%GFMs
\noindent Conversely, the GFMs perform better than the local models across evaluated levels. The fg\_LGBM achieved excellent performance with the best MASEs of 0.7732 and 0.8060 at the top and middle levels, respectively, while the nfg\_LGBM obtained the best MASE of 0.8235 at the bottom level. On average levels, comparisons between the loc\_LGBM and the global LightGBM indicate that nfg\_LGBM and fg\_LGBM outperformed the loc\_LGBM by 11.66\% and 21.42\%, respectively, and the fg\_LGBM-MINT obtained the best MASE of 0.8254. At the average of hierarchies, the nfg\_LGBM achieved 0.8418 MASE while the performance for fg\_LGBM has slightly rebounded with 0.8970 MASE. Although the two GFMs overwhelmingly outperform the local models, two exciting findings exist. One is that fg\_LGBM seems sensitive to relatively large values that reflect the best performances at the top and middle levels, 0.7732 and 0.8060, respectively. Meanwhile, the nfg\_LGBM focuses more on trends and changes at the bottom level.\\

%Variants
\noindent From the bottom to the top level, MASE is generally increasing, showing the deterioration of model performance. In \autoref{tab:1}, the performances of reconciled methods provide insights into how these methods influence these base forecasts. Compared with the performances of the original models, the TD method decreases the forecasting accuracy at the middle level, bottom level, average level and average hierarchies. For models except loc\_LGBM and the fg\_LGBM, the BU method usually augments the forecast accuracy at the top level, middle level, average level and average hierarchies. The BU method significantly mitigates the trend of deterioration for nfg\_LightGBM and only achieves a marginal effect for others. The MinT approach could maintain forecast accuracy at most levels and improve the performance of GFMs at the average level and the average hierarchies.

\begin{figure}
    \centering
    \includegraphics[width = 0.85\textwidth]{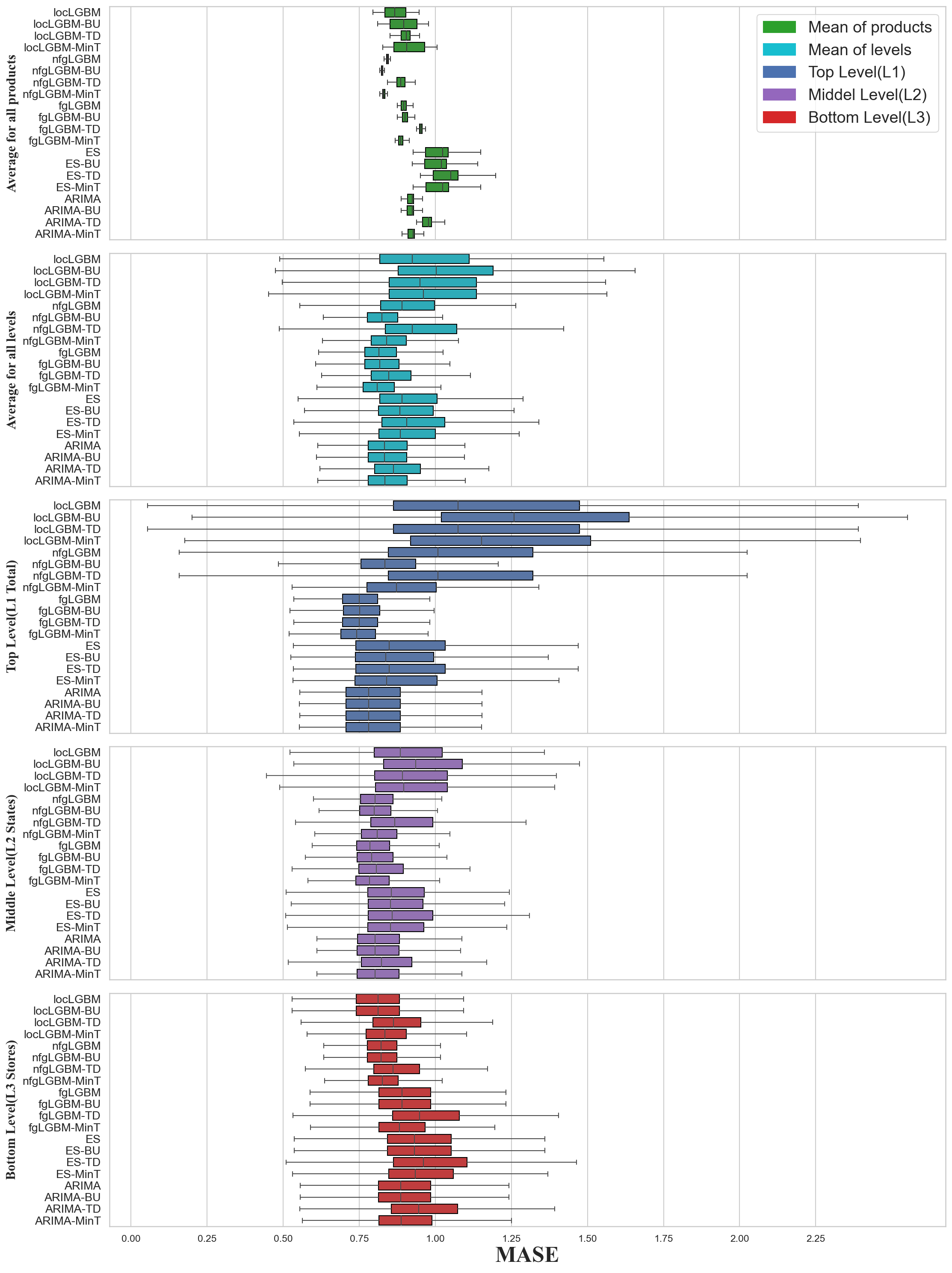}
    \caption{\textbf{MASEs for models and their variants}}
    \label{fig: Fig.5}
\end{figure}

\noindent The stability and reliability of forecasts are also crucial to decision-making. We looked at the performance further by investigating accuracy distributions. \autoref{fig: Fig.5} depicts boxplots of the MASE of the models across different levels. Compared with ARIMA and GFMs, the MASE of loc\_LGBM  is significantly dispersed at the evaluated levels except for the bottom level. At the top level, forecasts of local LightGBM distributed from around 0.05 to about 2.4, indicating that a considerable portion of forecasts is unreliable and the loc\_LGBM is vulnerable at the top level. The distribution of MASE generated by simple ES is only slightly tighter than the loc\_LGBM, and ARIMA achieved the most compact forecast distribution among local models.  Conversely, the forecasts of GFMs are more dense than the local models among five levels except for the nfg\_LGBM at the top level, showing that the forecasts of GFMs have higher reliability and stability, indicating the cross-hierarchies and cross-series information have a positive impact on not only the accuracy but also the reliability and stability of models.

\subsection{\normalfont\textbf{Statistical Test}}
\begin{figure}
    \centering
    \includegraphics[width = 0.89\textwidth]{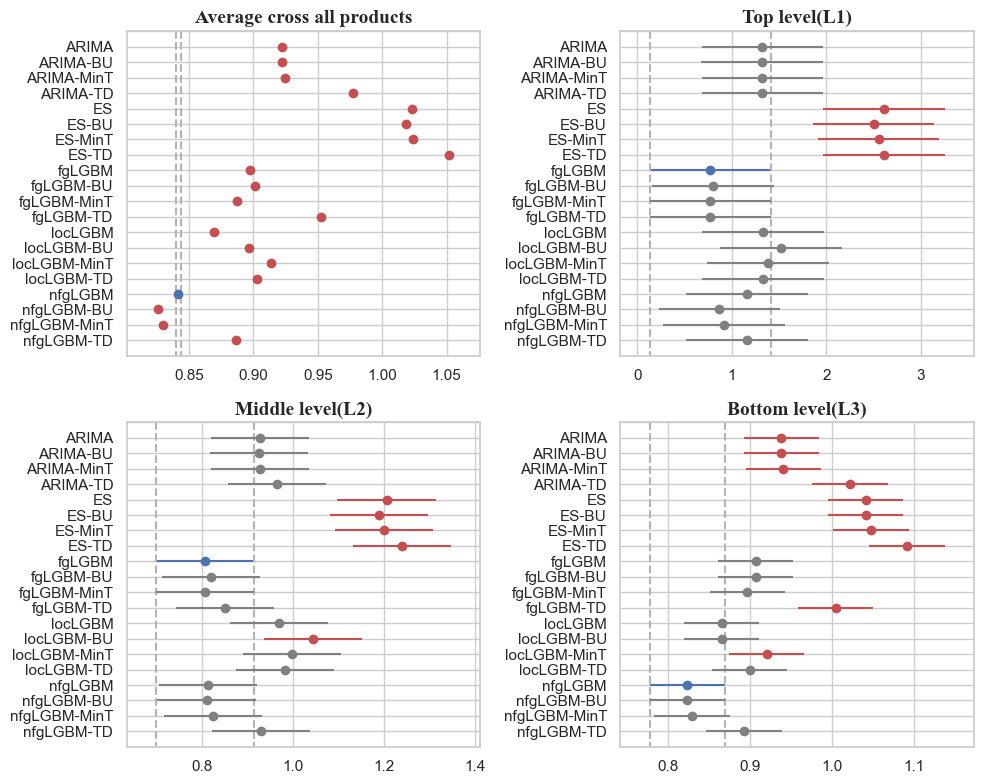}
    \caption{\textbf{MCB test for MASE of Models across Levels}}
    \label{fig: Fig.6}
\end{figure}

\noindent The Multiple Comparisons with the Best (MCB) tests were conducted to examine whether or not there was a statistically significant difference among the performances at the 95\% significance level. \autoref{fig: Fig.6} presents all MCB test results for models and their reconciled versions. At the lowest level and the average across all products, GFMs demonstrated distinction over local methods. However, at the middle and top levels, the performances of GFMs are close to local models. The nfg\_LGBM obtains the best performance at the lowest level, and its performance decreased as the levels increased, while the fg\_LGBM exhibited an improvement.

\section{CONCLUSION}\label{sec:conclusion}
\noindent The results indicate that various GFMs based on different information utilisation have advantages and disadvantages at distinct levels. Cross-series information extracts the same patterns among series in the same hierarchy, and the cross-hierarchies approach exploits all available patterns among series for hierarchies to improve forecasting. Although these two global models achieve more accurate forecasts than local models, they are only marginally better at the top and middle level than local models in terms of statistical significance.\\

\noindent In this study, we explored the effect caused by information utilisation on forecasting and proposed GFMs (nfg\_LGBM and fg\_LGBM). Our results show GFMs achieve higher accuracy and possess relatively lower model complexities compared to benchmarks and local models. The cross-hierarchies information augments the capacity of fg\_LGBM to forecast at the top and middle levels. The cross-information helps the nfg\_LGBM achieve more accurate forecasts at bottom levels. Simultaneously, the global modelling approach benefits models by generating more compact forecasts and increasing forecasting reliability. Regarding reconciliation methods, BU has a significant impact on accuracy improvement for the nfg\_LightGBM. The MinT considers accuracy among levels and mitigation of performance decay as the hierarchies and levels change, improving most models' accuracy.\\

\noindent The proposed methodology can benefit decision-makers in utilising the most relevant information from across different levels, providing insights on using useful information to generate reliable forecasts. Reconciled methods are still required to ensure the coherency of base forecasts. Maintaining the balance between accuracy of forecasts and cross-sectional learning across different levels should be concentrated in the future. Moreover, automatic information extraction that can generate coherent and accurate forecasts is another crucial work worth further investigation. Multiple variables, trends, seasonality, and other time series features that impact the performance of GFMs can be considered in future work to investigate the impact of information utilisation on performance in detail.

%\newpage

%\bibliographystyle{elsarticle-num-names} 
\bibliography{main.bib}

\begin{thebibliography}{18}
\expandafter\ifx\csname natexlab\endcsname\relax\def\natexlab#1{#1}\fi
\providecommand{\url}[1]{\texttt{#1}}
\providecommand{\href}[2]{#2}
\providecommand{\path}[1]{#1}
\providecommand{\DOIprefix}{doi:}
\providecommand{\ArXivprefix}{arXiv:}
\providecommand{\URLprefix}{URL: }
\providecommand{\Pubmedprefix}{pmid:}
\providecommand{\doi}[1]{\href{http://dx.doi.org/#1}{\path{#1}}}
\providecommand{\Pubmed}[1]{\href{pmid:#1}{\path{#1}}}
\providecommand{\bibinfo}[2]{#2}
\ifx\xfnm\relax \def\xfnm[#1]{\unskip,\space#1}\fi
%Type = Article
\bibitem[{Abolghasemi et~al.(2020)Abolghasemi, Beh, Tarr and Gerlach}]{abolghasemi2020demand}
\bibinfo{author}{Abolghasemi, M.}, \bibinfo{author}{Beh, E.}, \bibinfo{author}{Tarr, G.}, \bibinfo{author}{Gerlach, R.}, \bibinfo{year}{2020}.
\newblock \bibinfo{title}{Demand forecasting in supply chain: The impact of demand volatility in the presence of promotion}.
\newblock \bibinfo{journal}{Computers \& Industrial Engineering} \bibinfo{volume}{142}, \bibinfo{pages}{106380}.
%Type = Article
\bibitem[{Abolghasemi et~al.(2023)Abolghasemi, Rostami-Tabar and Syntetos}]{abolghasemi2023value}
\bibinfo{author}{Abolghasemi, M.}, \bibinfo{author}{Rostami-Tabar, B.}, \bibinfo{author}{Syntetos, A.}, \bibinfo{year}{2023}.
\newblock \bibinfo{title}{The value of point of sales information in upstream supply chain forecasting: an empirical investigation}.
\newblock \bibinfo{journal}{International Journal of Production Research} \bibinfo{volume}{61}, \bibinfo{pages}{2162--2177}.
%Type = Article
\bibitem[{Abolghasemi et~al.(2022)Abolghasemi, Tarr and Bergmeir}]{abolghasemi2022machine}
\bibinfo{author}{Abolghasemi, M.}, \bibinfo{author}{Tarr, G.}, \bibinfo{author}{Bergmeir, C.}, \bibinfo{year}{2022}.
\newblock \bibinfo{title}{Machine learning applications in hierarchical time series forecasting: Investigating the impact of promotions}.
\newblock \bibinfo{journal}{International Journal of Forecasting} \DOIprefix\doi{10.1016/j.ijforecast.2022.07.004}.
%Type = Article
\bibitem[{Bandara et~al.(2022)Bandara, Hewamalage, Godahewa and Gamakumara}]{BANDARA20221400}
\bibinfo{author}{Bandara, K.}, \bibinfo{author}{Hewamalage, H.}, \bibinfo{author}{Godahewa, R.}, \bibinfo{author}{Gamakumara, P.}, \bibinfo{year}{2022}.
\newblock \bibinfo{title}{A fast and scalable ensemble of global models with long memory and data partitioning for the m5 forecasting competition}.
\newblock \bibinfo{journal}{International Journal of Forecasting} \bibinfo{volume}{38}, \bibinfo{pages}{1400--1404}.
\newblock \DOIprefix\doi{doi.org/10.1016/j.ijforecast.2021.11.004}.
%Type = Incollection
\bibitem[{Duncan et~al.(2001)Duncan, Gorr and Szczypula}]{duncan2001forecasting}
\bibinfo{author}{Duncan, G.}, \bibinfo{author}{Gorr, W.}, \bibinfo{author}{Szczypula, J.}, \bibinfo{year}{2001}.
\newblock \bibinfo{title}{Forecasting analogous time series}, in: \bibinfo{editor}{Armstrong, J.} (Ed.), \bibinfo{booktitle}{Principles of Forecasting}. \bibinfo{publisher}{Springer}, \bibinfo{address}{Boston, MA}. volume~\bibinfo{volume}{30} of \textit{\bibinfo{series}{International Series in Operations Research \& Management Science}}, pp. \bibinfo{pages}{195--213}.
%Type = Article
\bibitem[{English and Abolghasemi(2024)}]{english2024improving}
\bibinfo{author}{English, L.}, \bibinfo{author}{Abolghasemi, M.}, \bibinfo{year}{2024}.
\newblock \bibinfo{title}{Improving the forecast accuracy of wind power by leveraging multiple hierarchical structure}.
\newblock \bibinfo{journal}{Sustainable Energy, Grids and Networks} , \bibinfo{pages}{101517}.
%Type = Article
\bibitem[{Holt(1957)}]{HOLT20045}
\bibinfo{author}{Holt, C.C.}, \bibinfo{year}{1957}.
\newblock \bibinfo{title}{Forecasting seasonals and trends by exponentially weighted moving averages}.
\newblock \bibinfo{journal}{International Journal of Forecasting} \bibinfo{volume}{20}, \bibinfo{pages}{5--10}.
\newblock \DOIprefix\doi{doi.org/10.1016/j.ijforecast.2003.09.015}.
%Type = Article
\bibitem[{Januschowski et~al.(2020)Januschowski, Gasthaus, Wang, Salinas, Flunkert, Bohlke-Schneider and Callot}]{januschowski2020criteria}
\bibinfo{author}{Januschowski, T.}, \bibinfo{author}{Gasthaus, J.}, \bibinfo{author}{Wang, Y.}, \bibinfo{author}{Salinas, D.}, \bibinfo{author}{Flunkert, V.}, \bibinfo{author}{Bohlke-Schneider, M.}, \bibinfo{author}{Callot, L.}, \bibinfo{year}{2020}.
\newblock \bibinfo{title}{Criteria for classifying forecasting methods}.
\newblock \bibinfo{journal}{International Journal of Forecasting} \bibinfo{volume}{36}, \bibinfo{pages}{167--177}.
\newblock \DOIprefix\doi{10.1016/j.ijforecast.2019.05.008}.
%Type = Inproceedings
\bibitem[{Ke et~al.(2017)Ke, Meng, Finley, Wang, Chen, Ma, Ye and Liu}]{ke2017lightgbm}
\bibinfo{author}{Ke, G.}, \bibinfo{author}{Meng, Q.}, \bibinfo{author}{Finley, T.}, \bibinfo{author}{Wang, T.}, \bibinfo{author}{Chen, W.}, \bibinfo{author}{Ma, W.}, \bibinfo{author}{Ye, Q.}, \bibinfo{author}{Liu, T.Y.}, \bibinfo{year}{2017}.
\newblock \bibinfo{title}{Lightgbm: A highly efficient gradient boosting decision tree}, in: \bibinfo{booktitle}{Advances in Neural Information Processing Systems (NIPS)}, pp. \bibinfo{pages}{3146--3154}.
%Type = Article
\bibitem[{Kourentzes et~al.(2014)Kourentzes, Petropoulos and Trapero}]{kourentzes2014improving}
\bibinfo{author}{Kourentzes, N.}, \bibinfo{author}{Petropoulos, F.}, \bibinfo{author}{Trapero, J.R.}, \bibinfo{year}{2014}.
\newblock \bibinfo{title}{Improving forecasting by estimating time series structural components across multiple frequencies}.
\newblock \bibinfo{journal}{International Journal of Forecasting} \bibinfo{volume}{30}, \bibinfo{pages}{291--302}.
%Type = Article
\bibitem[{Makridakis et~al.(2018)Makridakis, Spiliotis and Assimakopoulos}]{makridakis2018statistical}
\bibinfo{author}{Makridakis, S.}, \bibinfo{author}{Spiliotis, E.}, \bibinfo{author}{Assimakopoulos, V.}, \bibinfo{year}{2018}.
\newblock \bibinfo{title}{Statistical and machine learning forecasting methods: Concerns and ways forward}.
\newblock \bibinfo{journal}{PLoS ONE} \bibinfo{volume}{13}, \bibinfo{pages}{e0194889}.
\newblock \DOIprefix\doi{10.1371/journal.pone.0194889}.
%Type = Article
\bibitem[{Makridakis et~al.(2022)Makridakis, Spiliotis and Assimakopoulos}]{makridakis2021m5}
\bibinfo{author}{Makridakis, S.}, \bibinfo{author}{Spiliotis, E.}, \bibinfo{author}{Assimakopoulos, V.}, \bibinfo{year}{2022}.
\newblock \bibinfo{title}{The m5 competition: Background, organization, and implementation}.
\newblock \bibinfo{journal}{International Journal of Forecasting} \bibinfo{volume}{38}, \bibinfo{pages}{1325--1336}.
\newblock \DOIprefix\doi{10.1016/j.ijforecast.2021.07.007}.
%Type = Article
\bibitem[{Mejia et~al.(2021)Mejia, Avelar-Sosa, Mederos, Ramírez and {Díaz Roman}}]{MEJIA2021107371}
\bibinfo{author}{Mejia, J.}, \bibinfo{author}{Avelar-Sosa, L.}, \bibinfo{author}{Mederos, B.}, \bibinfo{author}{Ramírez, E.S.}, \bibinfo{author}{{Díaz Roman}, J.D.}, \bibinfo{year}{2021}.
\newblock \bibinfo{title}{Prediction of time series using an analysis filter bank of lstm units}.
\newblock \bibinfo{journal}{Computers \& Industrial Engineering} \bibinfo{volume}{157}, \bibinfo{pages}{107371}.
\newblock \URLprefix \url{https://www.sciencedirect.com/science/article/pii/S0360835221002758}, \DOIprefix\doi{https://doi.org/10.1016/j.cie.2021.107371}.
%Type = Article
\bibitem[{Montero-Manso and Hyndman(2021)}]{MONTEROMANSO20211632}
\bibinfo{author}{Montero-Manso, P.}, \bibinfo{author}{Hyndman, R.J.}, \bibinfo{year}{2021}.
\newblock \bibinfo{title}{Principles and algorithms for forecasting groups of time series: Locality and globality}.
\newblock \bibinfo{journal}{International Journal of Forecasting} \bibinfo{volume}{37}, \bibinfo{pages}{1632--1653}.
\newblock \DOIprefix\doi{https://doi.org/10.1016/j.ijforecast.2021.03.004}.
%Type = Article
\bibitem[{Nystrup et~al.(2021)Nystrup, Lindström, Møller and Madsen}]{NYSTRUP20211127}
\bibinfo{author}{Nystrup, P.}, \bibinfo{author}{Lindström, E.}, \bibinfo{author}{Møller, J.K.}, \bibinfo{author}{Madsen, H.}, \bibinfo{year}{2021}.
\newblock \bibinfo{title}{Dimensionality reduction in forecasting with temporal hierarchies}.
\newblock \bibinfo{journal}{International Journal of Forecasting} \bibinfo{volume}{37}, \bibinfo{pages}{1127--1146}.
\newblock \DOIprefix\doi{https://doi.org/10.1016/j.ijforecast.2020.12.003}.
%Type = Article
\bibitem[{Sezer et~al.(2020)Sezer, Gudelek and Ozbayoglu}]{SEZER2020106181}
\bibinfo{author}{Sezer, O.B.}, \bibinfo{author}{Gudelek, M.U.}, \bibinfo{author}{Ozbayoglu, A.M.}, \bibinfo{year}{2020}.
\newblock \bibinfo{title}{Financial time series forecasting with deep learning : A systematic literature review: 2005–2019}.
\newblock \bibinfo{journal}{Applied Soft Computing} \bibinfo{volume}{90}, \bibinfo{pages}{106181}.
\newblock \URLprefix \url{https://www.sciencedirect.com/science/article/pii/S1568494620301216}, \DOIprefix\doi{https://doi.org/10.1016/j.asoc.2020.106181}.
%Type = Article
\bibitem[{Smyl(2019)}]{smyl2019hybrid}
\bibinfo{author}{Smyl, S.}, \bibinfo{year}{2019}.
\newblock \bibinfo{title}{A hybrid method of exponential smoothing and recurrent neural networks for time series forecasting}.
\newblock \bibinfo{journal}{International Journal of Forecasting} \DOIprefix\doi{10.1016/j.ijforecast.2019.03.017}.
%Type = Article
\bibitem[{Trapero et~al.(2015)Trapero, Kourentzes and Fildes}]{trapero2015identification}
\bibinfo{author}{Trapero, J.R.}, \bibinfo{author}{Kourentzes, N.}, \bibinfo{author}{Fildes, R.}, \bibinfo{year}{2015}.
\newblock \bibinfo{title}{On the identification of sales forecasting models in the presence of promotions}.
\newblock \bibinfo{journal}{Journal of the Operational Research Society} \bibinfo{volume}{66}, \bibinfo{pages}{299--307}.
\newblock \DOIprefix\doi{10.1057/jors.2013.174}.

\end{thebibliography}





% \end{thebibliography}
\end{document}